\documentclass[runningheads]{llncs}
\usepackage{graphicx}

\newcommand{\etal}{\textit{et al.}}
\usepackage{multirow}
\usepackage{subcaption}
\usepackage{booktabs}

\usepackage{tikz}
\usepackage{comment} 
\usepackage{amsmath,amssymb} 
\usepackage{color}

\begin{document}
\pagestyle{headings}
\mainmatter
\def\ECCVSubNumber{100}  

\title{Attract, Perturb, and Explore: Learning a Feature Alignment Network for Semi-supervised Domain Adaptation} 

\titlerunning{Learning Feature Alignment for Semi-supervised Domain Adaptation}
%
\author{Taekyung Kim\orcidID{0000-0001-7401-098X} \and
Changick Kim}
\authorrunning{Taekyung Kim and Changick Kim}
%
\institute{Korea Advanced Institute of Science and Technology, Daejeon, South Korea\\
\email{\{tkkim93, changick\}@kaist.ac.kr}}
\maketitle

\begin{abstract}
Although unsupervised domain adaptation methods have been widely adopted across several computer vision tasks, it is more desirable if we can exploit a few labeled data from new domains encountered in a real application.
The novel setting of the semi-supervised domain adaptation (SSDA) problem shares the challenges with the domain adaptation problem and the semi-supervised learning problem.
However, a recent study shows that conventional domain adaptation and semi-supervised learning methods often result in less effective or negative transfer in the SSDA problem.
In order to interpret the observation and address the SSDA problem, in this paper, we raise the intra-domain discrepancy issue within the target domain, which has never been discussed so far.
Then, we demonstrate that addressing the intra-domain discrepancy leads to the ultimate goal of the SSDA problem.
We propose an SSDA framework that aims to align features via alleviation of the intra-domain discrepancy.
Our framework mainly consists of three schemes, i.e., attraction, perturbation, and exploration.
First, the attraction scheme globally minimizes the intra-domain discrepancy within the target domain.
Second, we demonstrate the incompatibility of the conventional adversarial perturbation methods with SSDA. Then, we present a domain adaptive adversarial perturbation scheme, which perturbs the given target samples in a way that reduces the intra-domain discrepancy.
Finally, the exploration scheme locally aligns features in a class-wise manner complementary to the attraction scheme by selectively aligning unlabeled target features complementary to the perturbation scheme.
We conduct extensive experiments on domain adaptation benchmark datasets such as DomainNet, Office-Home, and Office. Our method achieves state-of-the-art performances on all datasets.
\end{abstract}

\begin{figure*}[t]
\begin{center}
\includegraphics[width=0.93\linewidth]{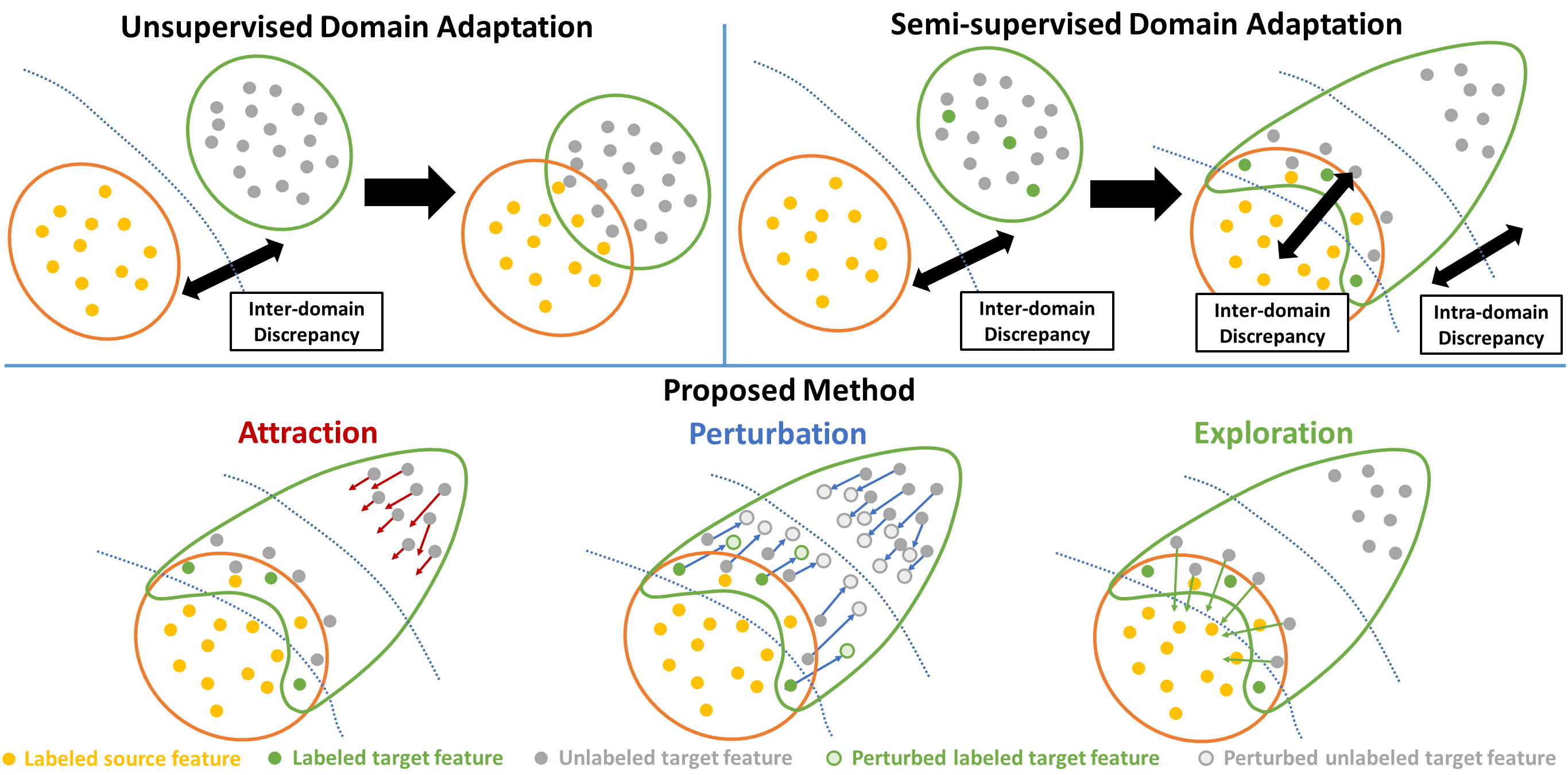}
\end{center}
\caption{
Conceptual descriptions of the feature alignment approaches. The top row describes the different feature alignment behaviors between the UDA and SSDA problem.
Supervision on labeled target samples attracts the corresponding features and their neighborhood toward the source feature cluster, which causes the intra-domain discrepancy.
The bottom row describes the proposed attraction, perturbation, and exploration schemes, which are explained in Section~\ref{sec:method} in detail.
}
\label{fig:concept}
\end{figure*}

\section{Introduction}

Despite the promising success of deep neural networks in several computer vision tasks, these networks often show performance degradation when tested beyond the training environment.
One way to mitigate this problem is to collect large amounts of data from the new domain and train the network.
Such heavy demands on data annotation cause great interest in domain adaptation and semi-supervised learning on deep neural networks.
However, most recent studies on deep domain adaptation are focused on unsupervised approaches, and deep semi-supervised learning is still concentrated on addressing the identical domain problem.
Though these methods can be directly applied to the semi-supervised domain adaptation (SSDA) problem only with an additional supervision on the extra labeled samples, a recent study~\cite{saito2019semi} reveals that unsupervised domain adaptation (UDA) methods and semi-supervised learning (SSL) methods often show less effective or even worse performances than just training on the labeled source and target samples in the SSDA problem.

In this paper, we introduce a new concept called \textit{intra-domain discrepancy} to analyze the failure of the UDA and SSL methods and address the SSDA problems.
Intra-domain discrepancy is a chronic issue in the SSDA problem that occurs during labeled sample supervision, but has never been discussed so far.
In the UDA problem, supervision on the labeled source samples does not critically affect the target domain distribution in general but implicitly attracts some alignable target features similar to the source features.
Thus, aligning the source and target domains by reducing their inter-domain discrepancy is reasonable.
However, in the SSDA problem, supervision on the labeled target samples enforces the corresponding features and their neighborhood to be attracted toward source feature clusters, which guarantees partial alignment between two domain distributions.
Besides, unlabeled target samples that less correlate with the labeled target samples are less affected by the supervision and eventually remain unaligned (Top row in Fig.~\ref{fig:concept}).
Thus, the target domain distribution is separated into an aligned target subdistribution and an unaligned target subdistribution, causing the intra-domain discrepancy within the target domain.
The failure of the UDA and SSL methods will be discussed in Section~\ref{sec:intra-domain} in detail.

Motivated by the insight, we propose an SSDA framework that aligns cross-domain features by addressing the intra-domain discrepancy within the target domain. 
Our framework focuses on enhancing the discriminability on the unaligned target samples and modulating the class prototypes, the representative features of each class.
It consists of three schemes, i.e., attraction, perturbation, and exploration, as shown in Fig.~\ref{fig:concept}.
First, the attraction scheme aligns the unaligned target subdistribution to the aligned target subdistribution through the intra-domain discrepancy minimization.
Second, we discuss why conventional adversarial perturbation methods are ineffective in the SSDA problem.
Unlike these approaches, our perturbation scheme perturbs target subdistributions into their intermediate region to propagate labels to the unaligned target subdistribution.
Note that our perturbation scheme does not ruin the already aligned target features since it additionally generates perturbed features temporarily for regularization.  
Finally, the exploration scheme locally modulates the prototypes in a class-aware manner complementary to the attraction and perturbation schemes.
We perform extensive experiments to evaluate the proposed method on domain adaptation datasets such as DomainNet, Office-Home, Office, and achieved state-of-the-art performances. 
We also deeply analyze our methods in detail.

Our contributions can be summarized as follows:
\begin{itemize} 
    \item We introduce the intra-domain discrepancy issue within the target domain in the SSDA problem.
    \item We propose an SSDA framework that addresses the intra-domain discrepancy issues via three schemes, i.e., attraction, perturbation, and exploration. 
    \begin{itemize}
    \item The attraction scheme aligns the unaligned target subdistribution to the aligned target subdistribution through the intra-domain discrepancy minimization.
    \item The perturbation scheme perturb target subdistributions into their intermediate region to propagate labels to the unaligned target subdistribution.
    \item The exploration scheme locally modulate the prototypes in a class-aware manner complementary to the attraction and perturbation schemes.   
    \end{itemize}
    \item We conduct extensive experiments on DomainNet, Office-Home, and Office. We achieve state-of-the-art performances among various methods, including vanilla deep neural networks, UDA, SSL, and SSDA methods.
\end{itemize}

\section{Related Work}
\subsection{Unsupervised Domain Adaptation}
The recent success of deep learning-based approaches and the following enormous demand for massive amounts of data attract great interest in domain adaptation (DA).
Even in the midst of significant interest, most recent works are focused on unsupervised domain adaptation (UDA).
Recent UDA methods can be categorized into three approaches.
The first approach is to reduce the cross-domain divergence. 
This can be achieved by minimizing the estimated domain divergence such as MMD~\cite{gretton2012kernel} or assimilating feature distributions through adversarial confusion using a domain classifier~\cite{ganin2014unsupervised,long2015learning,long2018conditional,long2016unsupervised}.
The second approach is to translate the appearance of one domain into the opposite domain so that the translated data can be regarded as sampled from the opposite domain~\cite{isola2017image,hu2018duplex,hoffman2018cycada}.
The last approach is to consider the source domain as partially labeled data and utilize the semi-supervised learning schemes.
For example, Drop-to-Adapt~\cite{lee2019drop} exploits a virtual adversarial perturbation scheme~\cite{miyato2018virtual}.
Recently, these approaches are widely adopted across several computer vision tasks beyond image classification such as object detection~\cite{chen2018domain,saito2018strong,kim2019diversify}, semantic segmentation~\cite{hsuan2018learning,hong2018conditional}, person re-identification~\cite{zheng2019joint}, and even in depth estimation~\cite{zheng2018t2net}.

\subsection{Semi-supervised Learning}
Similar to domain adaptation (DA), semi-supervised learning (SSL) has also attracted great attention as a way to overcome the shortages of the labeled data.
The difference between DA and SSL is that domain adaptation assumes to deal with data sampled from two distributions with significant domain discrepancy, while SSL assumes to deal with the labeled and unlabeled data sampled from the identical distribution.
With the rise of the deep learning approaches, several methods have been recently proposed for deep SSL.
Some works add data augmentation and regularize the model by enforcing a consistency between the given and the augmented data~\cite{laine2016temporal,tarvainen2017mean}.
Miyato~\etal~\cite{miyato2018virtual} extend this scheme by adversarially searching the bounded and small perturbation which leads the model to the most unstable state.
Laine and Aila~\cite{laine2016temporal} ensemble the prediction of the model by averaging them throughout the training phase, while Targainen and Valpola~\cite{tarvainen2017mean} ensemble the parameter of the model itself.
Other few works use self-training schemes with a memory module or a regularization through convergence speed~\cite{Chen_2018_ECCV,cicek2018saas}.
Recently, Wang~\etal~\cite{wang2019semi} propose an augmentation distribution alignment approach to explicitly address the empirical distribution mismatch problem in semi-supervised learning.

\subsection{Semi-supervised Domain Adaptation}
Semi-supervised domain adaptation (SSDA) is an important task which bridges the well-organized source distribution toward target distribution via partially labeled target samples, while a few works have explored so far~\cite{ao2017fast,donahue2013semi,yao2015semi,saito2019semi}.
Donahue~\etal~\cite{donahue2013semi} address the domain discrepancy by optimizing the auxiliary constrains on the labeled data.
Yao~\etal~\cite{yao2015semi} learn a subspace that can reduce the data distribution mismatch.
Ao~\etal~\cite{ao2017fast} estimate the soft label of the given labeled target sample with the source model and interpolated with the hard label for target model supervision.
Saito~\etal~\cite{saito2019semi} minimize the distance between the unlabeled target samples and the class prototypes through minimax training on entropy.
However, none of these methods discuss and mitigate the intra-domain discrepancy issue in the SSDA problem.
Different from previous works, we address the SSDA problem with a new perspective of the intra-domain discrepancy.

\begin{figure*}[t]
    \centering
    \begin{subfigure}[b]{0.31\textwidth}
        \includegraphics[width=\textwidth]{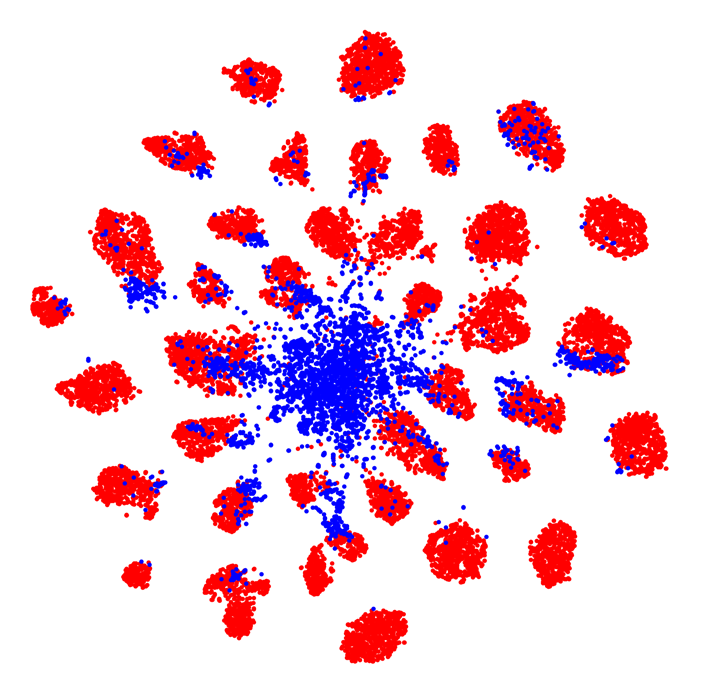}
        \caption{}
    \end{subfigure}
    \quad\quad
    \begin{subfigure}[b]{0.31\textwidth}
        \includegraphics[width=\textwidth]{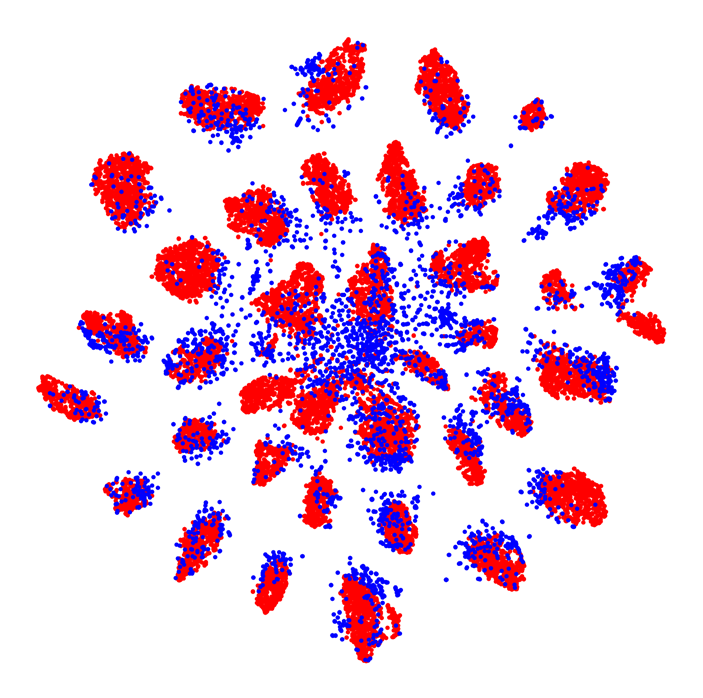}
        \caption{}
    \end{subfigure}

    \begin{subfigure}[b]{0.23\textwidth}
        \includegraphics[width=\textwidth]{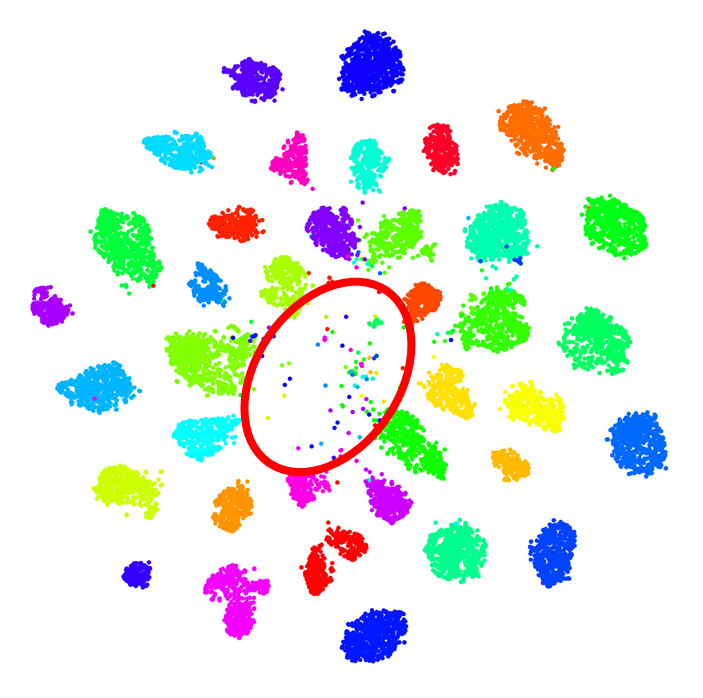}
        \caption{}
    \end{subfigure}
    \begin{subfigure}[b]{0.23\textwidth}
        \includegraphics[width=\textwidth]{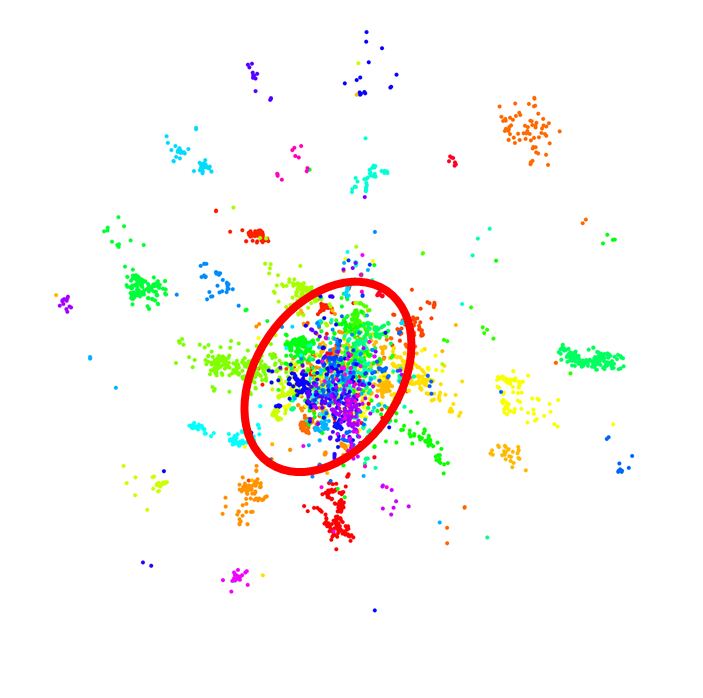}
        \caption{}
    \end{subfigure}
    \begin{subfigure}[b]{0.23\textwidth}
        \includegraphics[width=\textwidth]{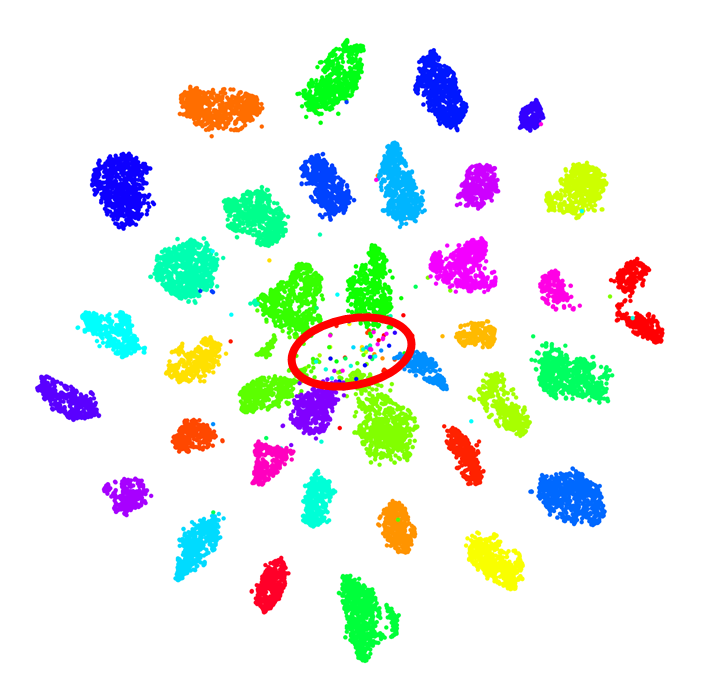}
        \caption{}
    \end{subfigure}
      \begin{subfigure}[b]{0.23\textwidth}
        \includegraphics[width=\textwidth]{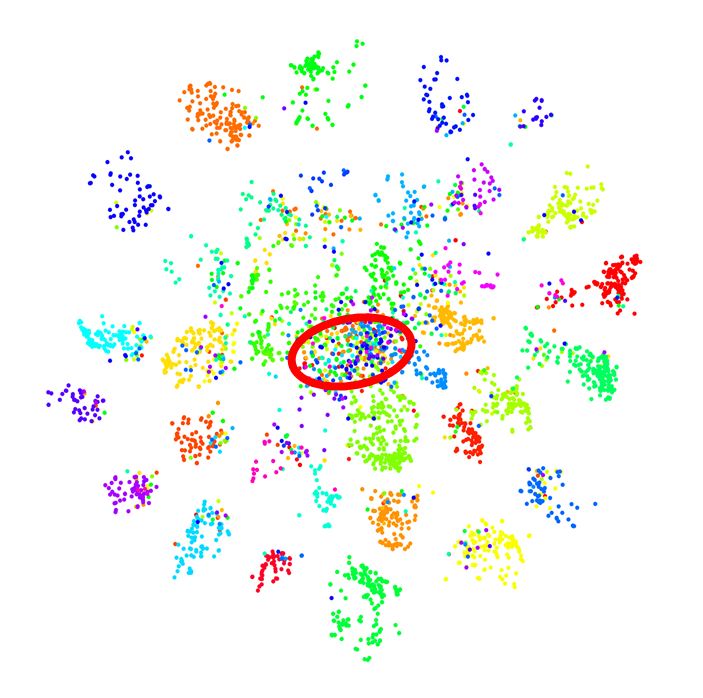}
        \caption{}
    \end{subfigure}
    
    \caption{ (a)-(f) The t-SNE visualization of the source and target features in (a) the UDA problem and (b) the SSDA problem with three target labels for each class. 
    We adopted the Real to Sketch scenario of the DomainNet dataset on the AlexNet backbone and visualized for partial classes.
    (c) and (e) visualize the source distribution of the UDA and SSDA problems and (d) and (f) visualize the target distribution of the UDA and SSDA problems, respectively.
    Even only three labeled target samples per class can attract their neighborhoods in this degree and separate the target domain into the aligned and unaligned subdistributions.
    }
    \label{fig:intra-domain}
\end{figure*}

\section{Intra-domain Discrepancy}\label{sec:intra-domain}
Intra-domain discrepancy of a domain is an internal distribution gap among subdistributions within the domain. 
Though we demonstrate the intra-domain discrepancy issue in the semi-supervised domain adaptation (SSDA) problem, such subdistributions can also appear in the unsupervised domain adaptation (UDA) problem since there usually exist target samples alignable to the source clusters.
However, since each domain generally has a unique correlation among domain samples, the target domain distribution is not easily separated into distinctive subdistributions, which eventually causes the insufficient intra-domain discrepancy.
Thus, the conventional inter-domain discrepancy minimization approaches have been effectively applied to the UDA problem.
In contrast, in the SSDA problem, supervision on the labeled target samples enforces the target domain to be separated into the aligned subdistribution and the unaligned subdistribution deterministically.
More specifically, as shown in the top row of Fig. 1, the presence of the label pulls the target samples and its neighborhoods toward source feature clusters of each corresponding labels. 
Besides, the unlabeled target samples which less correlate with the given labeled target samples are still located distant from source feature clusters, producing inaccurate and even meaningless inference results.
Figure~\ref{fig:intra-domain} demonstrates the existence of the intra-domain discrepancy within the target domain.
Though only three target labels per class are given, significant number of target samples are aligned while wrongly predicted target samples (red circle in Fig.~\ref{fig:intra-domain} (f)) are still located far from the source domain.

The presence of the intra-domain discrepancy makes the conventional domain adaptation methods less suitable for the SSDA problem.
The ultimate goal of domain adaptation is to enhance the discriminability on the target domain, and most of the error occurs on the unaligned target subdistribution in this case.
Thus, solving SSDA problems depends on how far the unaligned subdistribution is aligned.
However, common domain adaptation methods focus on reducing the inter-domain discrepancy between the source and target domains regardless of the intra-domain discrepancy within the target domain.
Since the existence of the aligned target subdistribution cause underestimation of the inter-domain discrepancy, the inter-domain discrepancy reduction approaches work less effectively in the SSDA problems.
Moreover, since the aligned target subdistribution is aligned in a class-aware manner, such approaches can even negatively affect.

Similarly, conventional semi-supervised learning (SSL) methods also suffer from the intra-domain discrepancy issue in the SSDA problem.
It stems from the different assumptions between the SSDA and SSL problems.
Since the SSL problem assumes to sample labeled and unlabeled data from the identical distribution, SSL methods mainly focus on propagating the correct labels to their neighbors.
In contrast, SSDA problems assume that there is a significant distribution divergence between the source and target domains, and that labeled samples are dominated by the source domain.
Since correctly predicted target samples are highly aligned with the source distribution, whereas incorrectly predicted target samples are located far from them, we can no longer assume that these target samples share the same distribution.
Thus, the SSL methods only propagate errors within the wrongly predicted subdistribution, and the propagation is also meaningless in the correctly predicted subdistribution due to the rich distribution of the source domain.
Motivated by the interpretation, we propose a framework that addresses the intra-domain discrepancy.

\section{Method}\label{sec:method}

\subsection{Problem Formulation}
Let us denote the set of source domain samples by $\mathcal{D}_{s}=\left\{\left(\mathbf{x}_{i}^{s}, {y_{i}}^{s}\right)\right\}_{i=1}^{m_{s}}$.
For the target domain, $\mathcal{D}_{t}=\left\{\left(\mathbf{x}_{i}^{t}, {y_{i}}^{t}\right)\right\}_{i=1}^{m_{t}}$ and $\mathcal{D}_{u} = \left\{\left(\mathbf{x}_{i}^{u} \right)\right\}_{i=1}^{m_{u}}$ denote the sets of labeled and unlabeled target samples, respectively. SSDA aims to enhance the target domain discriminability through training on $\mathcal{D}_{s}$, $\mathcal{D}_{t}$, and $\mathcal{D}_{u}$.

\subsection{Spherical Feature Space with Prototypes}
When aligning feature distributions, it is crucial to determine which feature space to adapt.
Even if the same method is used, performance may not be improved depending on the feature space applied.
Thus, we adopt similarity-based prototypical classifier in \cite{chen2019closer} to prepare suitable feature space for better adaptation.
Briefly, the prototypical classifier inputs normalized feature and compare the similarities among all class-wise prototypes, which reduces intra-class variations as results.
For the classifier training, we use a cross-entropy loss as our classification loss to train an embedding function $f_{\theta}(\cdot)$ with parameters $\theta$ and the prototypes $\mathbf{p}_k$ ($k$ = 1, ..., K) on the source domain samples and the labeled target samples:

\begin{equation}\label{loss_cls}
\begin{aligned}  
\mathcal{L}_{cls}& = \mathbb{E}_{(\mathbf{x}, y) \in \mathcal{D}_{s}
\cup\mathcal{D}_{t}} [ -\log p(y|\mathbf{x}, \mathbf{p})]\\
&= \mathbb{E}_{(\mathbf{x}, y) \in \mathcal{D}_{s}
\cup\mathcal{D}_{t}} \bigg[ - \log \Big(\frac{exp(\mathbf{p}_y\cdot f_{\theta}(\mathbf{x}) / T)}{\Sigma_{i=1}^{K}exp(\mathbf{p}_i\cdot f_{\theta}(\mathbf{x}) / T)} \Big) \bigg].
\end{aligned}
\end{equation}
While the prototypical classifier is trying to reduce the intra-class variation of the labeled sample features, the proposed schemes focus on aligning distributions of the normalized features on the spherical feature space.

\begin{figure*}[t]
\begin{center}
\includegraphics[width=\linewidth]{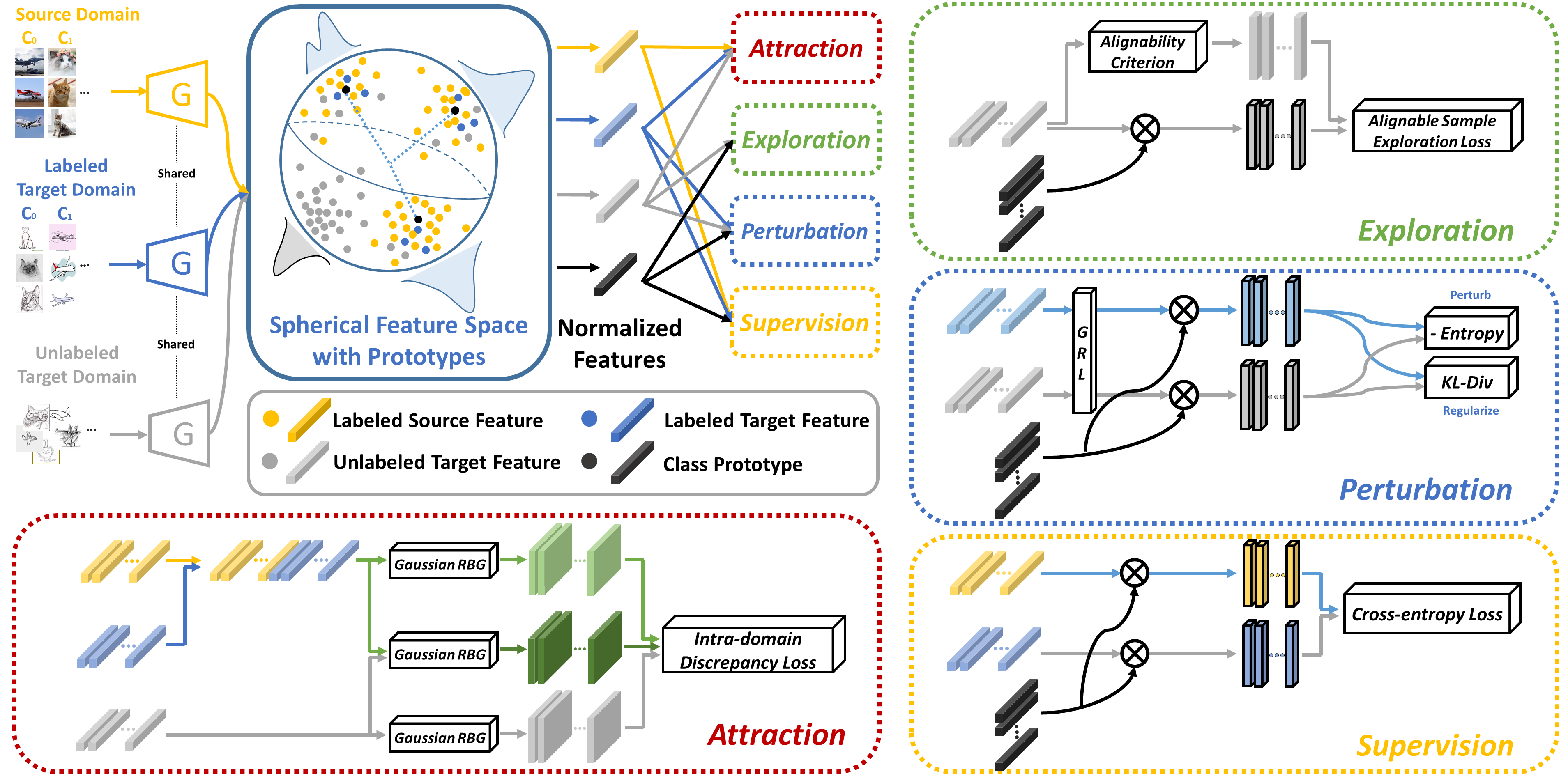}
\end{center}
  \caption{
    An overall framework of the proposed method. Our framework consists of the feature extractor, trainable class prototypes, supervision module, and each module for the proposed schemes. The class prototypes and all normalized features of the samples are embedded in the same spherical feature space.
  }
\label{fig:framework}
\end{figure*}

\subsection{Attraction Scheme}

The attraction scheme aims to globally align the unaligned target subdistribution to the aligned target subdistribution in a subdistribution-level through the estimated intra-domain discrepancy minimization.
The scheme measures the feature distribution divergence between the target subdistributions to estimate the intra-domain discrepancy within the target domain.
However, the limited number of the labeled target samples would not be sufficient to represent the feature distribution of the aligned target subdistribution.
Thus, motivated by the observation that the features of the aligned target subdistribution are highly aligned with that of the source domain in a class-aware manner, we instead use the complex distribution of the labeled source and target data.
For the empirical estimation of the intra-domain discrepancy, we adopt Maximum Mean Discrepancy (MMD)~\cite{gretton2012kernel}, a kernel two-sample test that measures the distribution difference.
We exploit a mixture $k(\cdot,\cdot)$ of Gaussian Radial Basis Function (RBF) kernels with multiple kernel widths $\sigma_{i}$ (i=1, ..., N).
Thus, the estimated intra-domain discrepancy on the spherical feature space can be written as:
\begin{equation}
\begin{aligned}
d(\mathcal{D}_{s}\cup \mathcal{D}_{t},\mathcal{D}_{u}) &= \mathbb{E}_{(\mathbf{x}, y), (\mathbf{x'}, y') \in \mathcal{D}_{s}\cup\mathcal{D}_{t}}[k(f_\theta(\mathbf{x}), f_\theta(\mathbf{x'}))] \\
&+ \mathbb{E}_{(\mathbf{z}, w), (\mathbf{z'}, w') \in \mathcal{D}_{u}}[k(f_\theta(\mathbf{z}), f_\theta(\mathbf{z'}))]\\
&- 2\mathbb{E}_{(\mathbf{x}, y) \in \mathcal{D}_{s}\cup\mathcal{D}_{t}, (\mathbf{z}, w)\in\mathcal{D}_{u}}[k(f_\theta(\mathbf{x}), f_\theta(\mathbf{z}))],
\end{aligned}
\end{equation} 
where $\mathbf{x'}$, $\mathbf{z}$, and $\mathbf{z'}$ represent samples, $y'$, $w$, and $w'$ represent the corresponding labels.
Since our attraction scheme directly minimizes the intra-domain discrepancy, the attraction loss can be written by:
\begin{equation}
\begin{aligned}
L_a = d(\mathcal{D}_{s}\cup \mathcal{D}_{t},\mathcal{D}_{u}).
\end{aligned}
\end{equation}

\subsection{Perturbation scheme}
Conventional adversarial perturbation, one of the semi-supervised learning (SSL) approaches, turns out to be ineffective or even cause negative transfer in the SSDA problem.
In the same context as discussed in Section~\ref{sec:intra-domain}, the labeled target samples and its neighborhoods are aligned to the source domain separated from the inaccurate target samples, causing the intra-domain discrepancy.
Then, the aligned features already guaranteed its confidence by rich information of the source domain, while the unaligned features can only propagate inaccurate predictions.
Thus, the perturbation on both the aligned and unaligned target subdistribution are less meaningful.

Unlike the common adversarial perturbation approaches, our scheme perturbs target subdistributions toward their intermediate region for 1) accurate prediction propagation from the aligned subdistribution to the unaligned subdistribution and 2) class prototypes modulation toward the region.
Such perturbation can be achieved by searching the direction of the anisotropically high entropy of the target features since the element-wise entropy increases as the feature move away from the prototype while the feature far from the prototypes can be attracted toward the prototypes.
Note that the perturbation scheme does not ruin the already aligned subdistribution since it temporally generates additional perturbed features of the aligned feature for regularization.
To achieve this, we first perturb the class prototypes in an entropy maximization direction.
Then, we optimize a small and bounded perturbation toward the perturbed prototypes.
Finally, we regularize the perturbed data and the given data through Kullback–Leibler divergence.
To summarize, the perturbation loss can be formulated as follows:
\begin{equation}\label{loss_usp}
\begin{aligned}
H_\mathbf{p}(\mathbf{x}) &= - \sum_{i=1}^{K} p(y=i|\mathbf{x}) \log p(y=i|\mathbf{x}, \mathbf{p})\\
r_{\mathbf{x}} &= \operatorname*{argmin}_{ \left\|r \right\| < \epsilon} \max_\mathbf{p} H_\mathbf{p}(\mathbf{x}+r)\\
\mathcal{L}_{p} &= \mathbb{E}_{\mathbf{x}\in\mathcal{D}_{u}}\bigg[\sum_{i=1}^{K}D_{KL}[p(y=i|\mathbf{x}, \mathbf{p}),p(y=i|\mathbf{x+r_x}, \mathbf{p})]\bigg]\\
 &+ \mathbb{E}_{(\mathbf{z}, w)\in\mathcal{D}_{t}} \bigg[\sum_{i=1}^{K}D_{KL}[p(y=i|\mathbf{z}, \mathbf{p}),p(y=i|\mathbf{z+r_z}, \mathbf{p})]\bigg].
\end{aligned}
\end{equation} 
where 
$H_\mathbf{p}(\cdot)$ is an element-wise entropy function defined upon similarities between the given feature and the prototypes,
$\mathbf{x}$ and $\mathbf{z}$ represent samples, and $y$ represent the corresponding label.

\subsection{Exploration scheme}
The exploration scheme aims to locally modulate the prototypes in a class-aware manner complementary to the attraction scheme, while selectively aligns the unlabeled target features via suitable criteria complementary to the perturbation scheme.
Though the attraction scheme globally aligns the target subdistributions on the feature space regardless of the prototypes, it does not explicitly enforce the prototypes to be modulated, which can be complemented by local and class-aware alignment.
On the other hand, since the perturbation scheme regularizes the perturbed features of the anisotropically high entropy, the entropy of the perturbed feature and its neighborhood gradually became low.
The exploration scheme aligns these features so that their entropy became isotropic, and thus the aligned features can be perturbed farther toward the unaligned subdistribution.
To practically achieve this, we selectively collect unlabeled target data with its element-wise entropy less than a certain threshold, then apply a cross-entropy loss with the class of the nearest prototype.
The objective function of the exploration scheme can be written as follows:

\begin{equation}
\begin{aligned}
M_\epsilon &= \{\mathbf{x}\in\mathcal{D}_{u} | H_\mathbf{p}(\mathbf{x}) < \epsilon\}\\
\hat{y}_{\mathbf{x}} &= \operatorname*{argmax}_{i\in\{1,...,K\}} p(y=i|\mathbf{x}, \mathbf{p})\\
\mathcal{L}_{e} &= \mathbb{E}_{\mathcal{D}_{u}}[-\mathbf{1}_{M_\epsilon}(\mathbf{x})\log p(y=\hat{y}_{\mathbf{x}}|\mathbf{x}, \mathbf{p})].
\end{aligned}
\end{equation} 
where 
$M_\epsilon$ is a set of unlabeled target data with entropy value less than a hyperparameter $\epsilon$, 
and $\mathbf{1}_{M_\epsilon}(\cdot)$ is an indicator function that filters out alignable samples from the given unlabeled target samples.

\subsection{Overall framework and training objective}
The overall training objective of our method is the weighted sum of the supervision loss, the attraction loss, the perturbation loss, and the exploration loss.
The optimization problem can be formulated as follows:
\begin{equation}
\begin{aligned}
\min_{\mathbf{p}, \theta} \mathcal{L}_{cls} + \alpha\mathcal{L}_{a} + \beta\mathcal{L}_{e} + \gamma\mathcal{L}_{p}.
\end{aligned}
\end{equation} 
We integrated all the schemes into one framework, as shown in the Fig.~\ref{fig:framework}.

\begin{table*}[t]
\caption{Classification accuracy ($\%$) on the DomainNet dataset on the AlexNet and ResNet-34 backbone networks. 
The performance comparisons were done for seven scenarios with one or three labeled target samples for each class. }
\begin{center}
\scalebox{0.74}{
\begin{tabular}{l|l|cccccccccccccc|cc}
\toprule[1.5pt]
 \multirow{2}{*}{Net} & \multirow{2}{*}{Method}       &\multicolumn{2}{c}{R to C}&\multicolumn{2}{c}{R to P} & \multicolumn{2}{c}{P to C}  & \multicolumn{2}{c}{C to S} & \multicolumn{2}{c}{S to P} & \multicolumn{2}{c}{R to S} & \multicolumn{2}{c}{P to R}     &\multicolumn{2}{|c}{MEAN} \\ 
& &1\scriptsize{-shot}&3\scriptsize{-shot} &1\scriptsize{-shot}&3\scriptsize{-shot}&1\scriptsize{-shot}&3\scriptsize{-shot} &1\scriptsize{-shot}&3\scriptsize{-shot}&1\scriptsize{-shot}&3\scriptsize{-shot}&1\scriptsize{-shot}&3\scriptsize{-shot} &1\scriptsize{-shot}&3\scriptsize{-shot} &1\scriptsize{-shot}&3\scriptsize{-shot}  \\ \hline
 \multirow{8}{*}{AlexNet} &S+T& 43.3   & 47.1 & 42.4   & 45.0 & 40.1   & 44.9 & 33.6   & 36.4 & 35.7   & 38.4 & 29.1 & 33.3 & 55.8   & 58.7 & 40.0 & 43.4 \\
&DANN & 43.3   & 46.1 & 41.6   & 43.8 & 39.1   & 41.0 & 35.9   & 36.5 &36.9   & 38.9 & 32.5 & 33.4 & 53.6   & 57.3 & 40.4 & 42.4 \\
&ADR & 43.1        & 46.2 &    41.4    & 44.4 &    39.3    & 43.6 & 32.8        &   36.4   &  33.1      &  38.9    &   29.1   &  32.4    & 55.9  & 57.3 & 39.2  & 42.7 \\
&CDAN& 46.3   & 46.8 & 45.7   & 45.0 & 38.3   & 42.3 & 27.5   & 29.5 & 30.2   & 33.7 & 28.8 & 31.3 & 56.7   & 58.7 & 39.1 & 41.0 \\
&ENT          & 37.0   & 45.5 & 35.6   & 42.6 & 26.8   & 40.4 & 18.9   & 31.1 & 15.1   & 29.6 & 18.0 & 29.6 & 52.2   & 60.0 & 29.1 & 39.8 \\
&MME & \bf{48.9}& \bf{55.6} & 48.0   & 49.0 & 46.7 &51.7 & 36.3   & 39.4 & \bf{39.4}   & \bf{43.0} & 33.3 & 37.9 & 56.8   & 60.7 & 44.2 & 48.2\\
&SagNet & 45.8 & 49.1 & 45.6 & 46.7 & 42.7 & 46.3 & 36.1 & 39.4 & 37.1 & 39.8 & 34.2 & 37.5 & 54.0 & 57.0 & 42.2 & 45.1\\
&Ours & 47.7& 54.6 & \bf{49.0}   & \bf{50.5} & \bf{46.9}   &\bf{52.1} & \bf{38.5}   &\bf{42.6} & 38.5   & 42.2 &\bf{33.8} &\bf{38.7} & \bf{57.5}   & \bf{61.4} &\bf{44.6} & \bf{48.9}\\\hline

\multirow{8}{*}{ResNet} &S+T    & 55.6 & 60.0   & 60.6 & 62.2   & 56.8 & 59.4   & 50.8 & 55.0   & 56.0 & 59.5 & 46.3 & 50.1   & 71.8 & 73.9 & 56.9 & 60.0 \\
&DANN   & 58.2 & 59.8   & 61.4 & 62.8   & 56.3 & 59.6   & 52.8 & 55.4   & 57.4 & 59.9 & 52.2 & 54.9   & 70.3 & 72.2 & 58.4 & 60.7 \\
&ADR    & 57.1 & 60.7 & 61.3 & 61.9 & 57.0 & 60.7 & 51.0 & 54.4 & 56.0 & 59.9 & 49.0 & 51.1 & 72.0 & 74.2 & 57.6 & 60.4
  \\
&CDAN   & 65.0 & 69.0   & 64.9 & 67.3   & 63.7 & 68.4   & 53.1 & 57.8   & 63.4 & 65.3 & 54.5 & 59.0   & 73.2 & 78.5 & 62.5 & 66.5 \\
&ENT    & 65.2 & 71.0   & 65.9 & 69.2   & 65.4 & 71.1   & 54.6 & 60.0   & 59.7 & 62.1 & 52.1 & 61.1   & 75.0 & 78.6 & 62.6 & 67.6 \\
&MME   & 70.0 & 72.2   & 67.7 & 69.7   & 69.0 & 71.7   & 56.3 & 61.8   & \bf{64.8} & \bf{66.8} & 61.0 & 61.9   & 76.1 & 78.5 & 66.4 & 68.9\\
&SagNet & 59.4 & 62.0 & 61.9 & 62.9 & 59.1 & 61.5 & 54.0 & 57.1 & 56.6 & 59.0 & 49.7 & 54.4 & 72.2 & 73.4 & 59.0 & 61.5\\
&Ours & \bf{70.4}& \bf{76.6} & \bf{70.8}   & \bf{72.1} & \bf{72.9}  &\bf{76.7} & \bf{56.7}   & \bf{63.1} & 64.5  & 66.1 & \bf{63.0} & \bf{67.8} &\bf{76.6}   & \bf{79.4} & \bf{67.6} &\bf{71.7}\\
\bottomrule[1.5pt]
\end{tabular}}
\end{center}
\label{tb:result_domainet}
\end{table*}

\section{Experiments}
\subsection{Experimental Setup}
\noindent{\bf Datasets.} {\bf DomainNet}~\cite{peng2019moment} is a recently released large-scale domain adaptation benchmark dataset that contains six domains and approximately 0.6 million images with 345 classes.
{\bf Office-Home}~\cite{venkateswara2017Deep} and {\bf Office}~\cite{office} are standard benchmarks for domain adaptation.
Office-Home consists of Art, Clipart, Product, and Real-world domain with 65 classes.
Office consists of Amazon, Webcam, and DSLR domains with 31 classes.

\noindent{\bf Evaluation tasks.}
For a fair comparison with the state-of-the-art SSDA method ~\cite{saito2019semi}, we performed experiments on 7 adaptation scenarios on the four domains (Real, Clipart, Painting, Sketch) with 126 classes for DomainNet, 12 adaptation scenarios on all the domains for Office-Home, and two challenging adaptation scenarios on  Office. 
One or three labeled target samples are given for each class for these scenarios.
Additionally, we compared the performances for 5, 10 and 20 labeled target samples for each class. 

\noindent{\bf Implementation details.}
We adopted AlexNet and ResNet-34 for the backbone network.
Every mini-batch consists of the same number of labeled source and target samples with a doubled number of unlabeled target samples.
We prepared 32 and 24 samples for each split of the mini-batch for AlexNet and Resnet-34, respectively.
We used the Stochastic Gradient Descent (SGD) optimizer with an initial learning rate of 0.01, a momentum of 0.9, and a weight decay of 0.0005.
All implementations were done in PyTorch~\cite{paszke2017automatic} and on a single GeForce Titan XP GPU.

\begin{table*}[t]
\caption{Classification accuracy ($\%$) on the Office-Home dataset with the AlexNet and ResNet-34 backbone networks. 
The performance comparisons were done for a total of 12 scenarios on three-shot setting.}
\begin{center}
\scalebox{0.74}{
\begin{tabular}{c|l|cccccccccccc|c}
\toprule[1.5pt] 
Net& Method       &R to C& R to P & R to A & P to R & P to C & P to A & A to P & A to C & A to R & C to R & C to A & C to P & MEAN \\\hline

\multirow{7}{*}{AlexNet}& S+T          & 44.6   & 66.7   & 47.7   & 57.8   & 44.4   & 36.1   & 57.6   & 38.8   & 57.0   & 54.3   & 37.5   & 57.9   & 50.0 \\
& DANN         & 47.2   & 66.7   & 46.6   & 58.1   & 44.4   & 36.1   & 57.2   & 39.8   & 56.6   & 54.3   & 38.6   & 57.9   & 50.3 \\
& ADR          & 45.0   & 66.2   & 46.9   & 57.3   & 38.9   & 36.3   & 57.5   & 40.0   & 57.8   & 53.4   & 37.3   & 57.7   & 49.5 \\
& CDAN         & 41.8   & 69.9   & 43.2   & 53.6   & 35.8   & 32.0   & 56.3   & 34.5   & 53.5   & 49.3   & 27.9   & 56.2   & 46.2 \\
& ENT          & 44.9   & 70.4   & 47.1   & 60.3   & 41.2   & 34.6   & 60.7   & 37.8   & 60.5   & 58.0   & 31.8   & 63.4   & 50.9 \\
& MME & 51.2   & 73.0   & 50.3   & \bf{61.6}   & 47.2   & 40.7   & 63.9   & 43.8   & \bf{61.4}   & \bf{59.9}  & \bf{44.7}  & 64.7   &  55.2 \\
& Ours & \bf{51.9}   & \bf{74.6}   & \bf{51.2}   & {\bf 61.6}   & \bf{47.9}   & \bf{42.1}  & \bf{65.5}   & \bf{44.5}   & 60.9   & 58.1  & 44.3 & \bf{64.8}   & \bf{55.6}\\\hline

\multirow{7}{*}{ResNet} & S+T          &  55.7  & 80.8 & 67.8 & 73.1 & 53.8 & 63.5 & 73.1 & 54.0 & 74.2 & 68.3 & 57.6 & 72.3 & 66.2\\
& DANN         &  57.3 & 75.5 & 65.2 & 69.2 & 51.8 & 56.6 & 68.3 & 54.7 & 73.8 & 67.1 & 55.1 & 67.5 & 63.5 \\
& CDAN         &  61.4 & 80.7 & 67.1 & 76.8 & 58.1 & 61.4 & 74.1 & 59.2 & 74.1 & 70.7 & 60.5 & 74.5 & 68.2 \\
& ENT          &  62.6 & 85.7 & 70.2 & 79.9 & 60.5 & 63.9 & 79.5 & 61.3 & 79.1 & 76.4 & 64.7 & 79.1 & 71.9\\
& MME &  64.6 & 85.5 & 71.3 & 80.1 & 64.6 & 65.5 & 79.0 & 63.6 & 79.7 & 76.6 & {\bf 67.2} & 79.3 & 73.1 \\
& Ours & {\bf 66.4} & {\bf 86.2} & {\bf 73.4} & {\bf 82.0} & {\bf 65.2} & {\bf 66.1} & {\bf 81.1} & {\bf 63.9} & {\bf 80.2} & {\bf 76.8} & 66.6 & {\bf 79.9} & \bf{74.0}\\
        \bottomrule[1.5pt]
\end{tabular}}
\end{center}
\label{tb:result_office-home}
\end{table*}

\noindent{\bf Baselines.}
We compared our method with the semi-supervised domain adaptation (SSDA), unsupervised domain adaptation (UDA), semi-supervised learning (SSL), and no adaptation methods.
More specifically, the baselines consist of {\bf MME}~\cite{saito2019semi}, {\bf SagNet}~\cite{nam2019reducing}, {\bf DANN}~\cite{ganin2014unsupervised}, {\bf ADR}~\cite{saito2018adversarial}, {\bf CDAN}~\cite{long2018conditional}, {\bf ENT}~\cite{grandvalet2005semi}, and non-adapted model.
For the UDA methods (DANN, ADR, and CDAN), the labeled target samples were supervised during the training process.
S+T is a vanilla model trained on all labeled samples.
DANN confuses the cross-domain distributions through adversarial learning.
ADR adopts the dropout scheme to modify the decision boundary for feature alignment.
CDAN adversarially aligns the feature by fooling the conditional domain discriminator.
ENT is an SSL method that minimizes the entropy of the unlabeled target data.
For the fair comparison, all the methods have the same backbone architecture with our method.

\subsection{Results}
\noindent{\bf Performance Comparison on DomainNet.}
We summarized the classification accuracies of 7 scenarios on the DomainNet dataset in Table~\ref{tb:result_domainet}.
On average, our method outperformed the best-performed baseline by 2.8$\%$ in the three-shot setting and 1.2$\%$ in the one-shot setting on ResNet-34, and by 0.7$\%$ in the three-shot setting and 0.4$\%$ in the one-shot setting on AlexNet.
Moreover, our method outperformed most of the cases except for a few adaptation tasks.
On the other hand, though UDA methods like DANN and ADR performed slightly better than S+T when only one labeled target per class is given, these methods become less effective or even cause negative transfer as the number of the labeled target samples increases.
It verifies our statement that conventional domain adaptation methods are often less beneficial than the partial alignment effect from the given target labels.
ENT showed significant improvement on ResNet-34, while it shows degenerative performance on AlexNet. Moreover, the performance enhancement gap increased as the number of labeled target samples increases, which will be discussed in more detail in Section~\ref{sec:analysis}.

\begin{table}[t]
\caption{
Classification accuracy ($\%$) on the Office dataset with three-shot setting.
}
\begin{center}
\scalebox{0.75}{
\begin{tabular}{c|l|cc|c}
\toprule[1.5pt]
Net & Method    &\multicolumn{1}{c}{W to A}&\multicolumn{1}{c|}{D to A} & \multicolumn{1}{c}{MEAN} \\\hline
\multirow{7}{*}{AlexNet}& S+T     & 61.2  & 62.4 & 61.8 \\
& DANN    & 64.4  & 65.2 & 64.8 \\
& ADR     & 61.2  & 61.4 & 61.3 \\
& CDAN    & 60.3  & 61.4 & 60.8  \\
& ENT     & 64.0  & 66.2 & 65.1 \\
& MME     &67.3    &   67.8 & 67.6 \\
& Ours    &  \bf{ 67.6 }& \bf{69.0} & \bf{68.3}  \\
\bottomrule[1.5pt]
\end{tabular}}
\end{center}
\label{tb:tb_office}
\end{table}

\begin{figure*}[t]
    \centering
    \begin{subfigure}[b]{0.43\textwidth}
        \includegraphics[width=\textwidth]{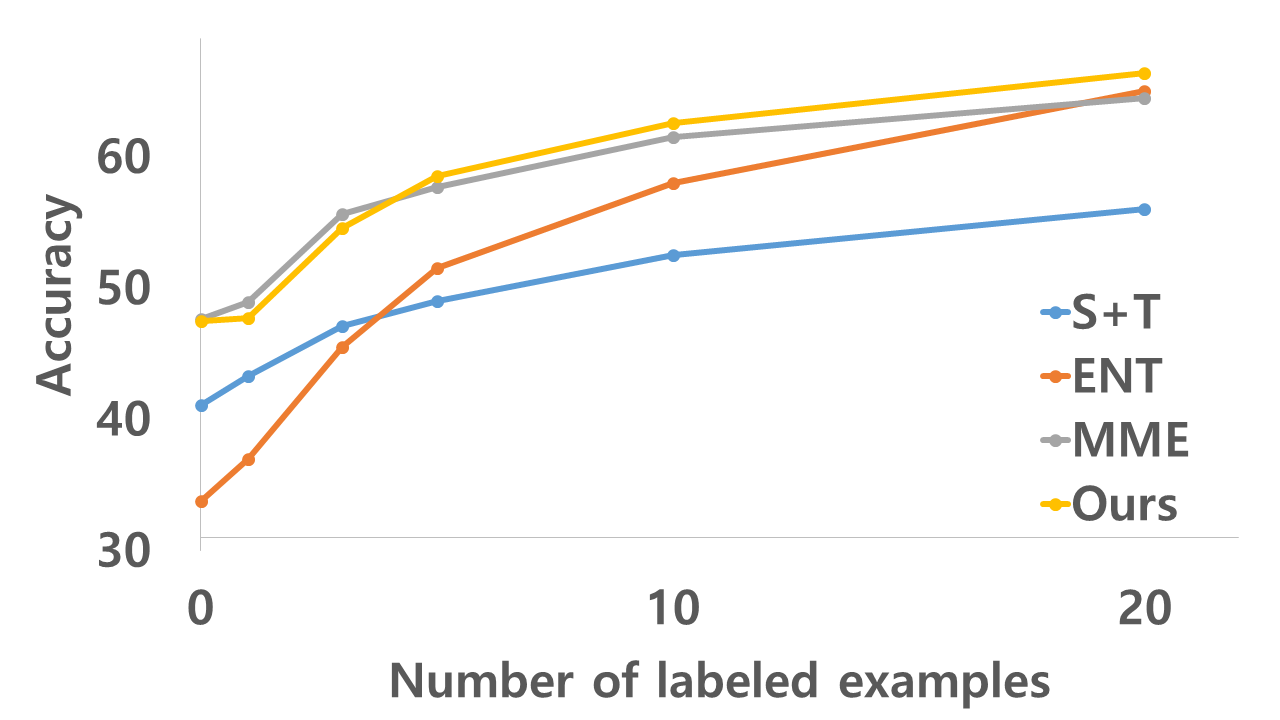}
        \caption{AlexNet}
    \end{subfigure}
    \begin{subfigure}[b]{0.43\textwidth}
        \includegraphics[width=\textwidth]{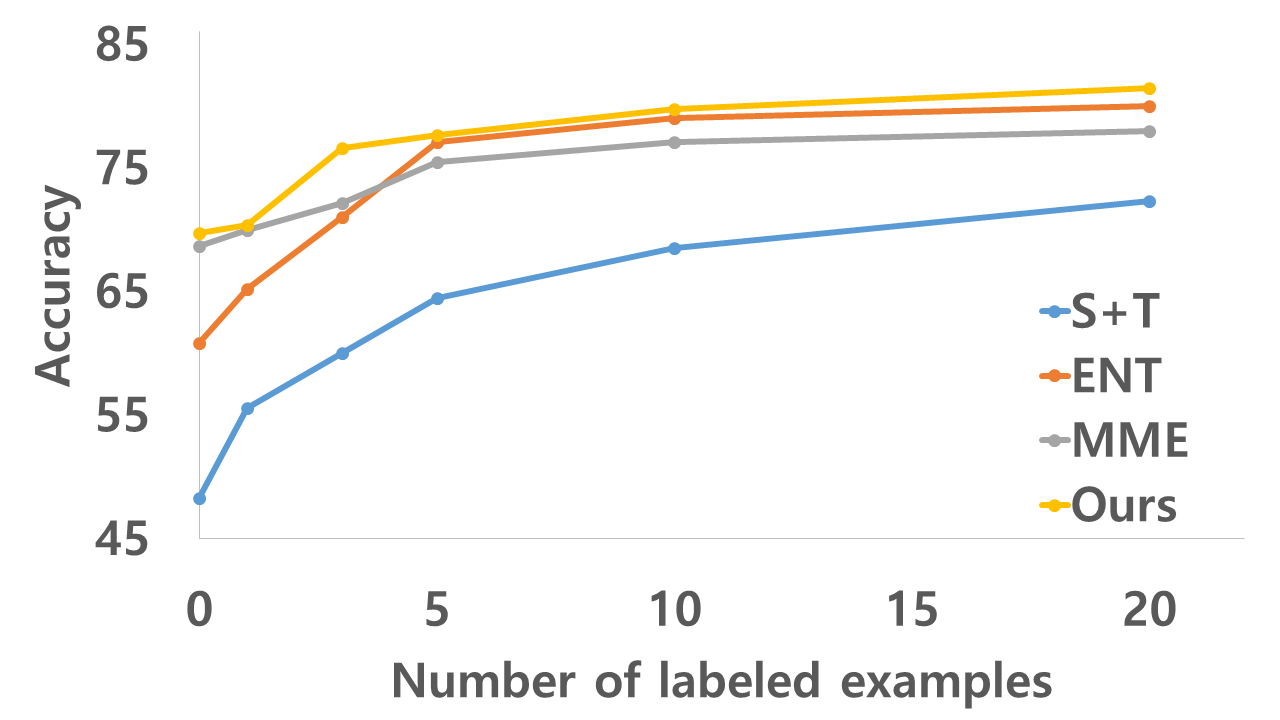}
        \caption{ResNet-34}
    \end{subfigure}
   
    \caption{Trend in classification accuracy ($\%$) with varying number of labeled target samples per class. The experiments are conducted on the Real to Clipart scenario of the DomainNet dataset.}
    \label{fig:varying}
\end{figure*}

\noindent{\bf Performance Comparison on Office-Home and Office.} The comparison results of our method with the baselines on the Office-Home dataset are reported in Table~\ref{tb:result_office-home}.
Our method outperformed all the baselines regardless of the backbone network on average.
Similar to DomainNet, our method showed the best performance in most of the scenarios.
While DANN performed at least similar to S+T on the AlexNet backbone, it showed degenerative performance on the ResNet-34 backbone.
It demonstrates that capacity difference of the backbone network causes the difference in the degree of the target label exploitation, and DANN performed less effective than the exploitation of ResNet-34.
The considerable results of ENT are reasonable since the three-shot setting provides approximately $5\sim10\%$ of the target labels for training, and such ratio is quite rich in a perspective of the SSL problem.
Table~\ref{tb:tb_office} showed the performance comparison on the Office dataset and our method also outperformed other baseline on the dataset.

\begin{table}[t]
\caption{Ablation study results of the proposed schemes on the Real to Sketch task of DomainNet with three-shot setting. 
}
\begin{center}
\scalebox{0.83}{
\begin{tabular}{c|l|ccc|ccccccc|c}
\toprule[1.5pt]
Net & Method & Attract & Explore & Perturb & R to C & R to P & P to C & C to S & S to P & R to S & P to R & MEAN\\ \hline
 \multirow{9}{*}{AlexNet} & S+T&&& & 47.1  & 45.0 & 44.9 & 36.4 & 38.4 & 33.3 & 58.7 & 43.4\\
 &       DANN &&& & 46.1 & 43.8 & 41.0 & 36.5 & 38.9 & 33.4 & 57.3 & 42.4\\
 &       MMD  &&& & 47.9 & 45.5 & 44.6 & 38.1 & 38.4 & 35.5 & 56.6 & 43.8      \\
 &       VAT  &&& & 46.1 & 43.8 & 44.3 & 35.6 & 38.2 & 31.8 & 57.7 & 42.5 \\\cline{2-13}
 &        \multirow{5}{*}{Ours} &\checkmark&& & 50.2 & 46.2 & 47.5 & 40.8 & 41.3 & 37.2 & 59.8 & 46.1\\
 &        &\checkmark&\checkmark&&  53.9 & 49.8 & 50.5 & 42.0 & 41.9 & 38.0 & 60.7 & 48.3  \\
 &        &&&\checkmark& 57.2 & 47.5 & 54.1 & 38.8 & 39.7 & 38.5 & 59.2 & 47.9\\
 &         &\checkmark&\checkmark&\checkmark& 54.6     & 50.5 & 52.1 & 42.6 & 42.2 & 38.7 & 61.4 & 48.9   \\
\bottomrule[1.5pt]
\end{tabular}}
\end{center}
\label{tb:analysis_ablation}
\end{table}

\subsection{Analysis}\label{sec:analysis}
\noindent{\bf Performance comparison with varying number of target labels.}
We compared the behavior of the methods by varying the number of labeled target samples from 0 to 20 for each class.
As shown in Fig.~\ref{fig:varying}, our methods showed superior performance for a large number of target labels even on the scenario where our method worked less effectively on a one-shot or three-shot setting.
Moreover, it outperformed the other baselines throughout all the cases on the ResNet-34 backbone.
On the other hand, ENT also significantly enhanced the accuracy for a large number of target labels, and it even outperformed the state-of-the-art SSDA methods when more than twenty and five target labels are given per class for the AlexNet and ResNet-34 backbone networks, respectively.
It is reasonable since the increase of the labeled target sample ratio assimilates the SSDA problem to the SSL problem, which is suitable for SSL methods.

\noindent{\bf Ablation Study on the proposed schemes.}
We conducted an ablation study on our schemes.
To verify the effectiveness, we additionally evaluated DANN, MMD, and VAT~\cite{miyato2018virtual} on the DomainNet dataset.
As shown in Table~\ref{tb:analysis_ablation}, DANN and MMD rarely worked or even caused negative transfer while our attraction scheme showed meaningful improvement on average.
It verifies that conventional UDA methods that focus on reducing the inter-domain discrepancy suffer from the intra-domain discrepancy issue, and it can be addressed by the intra-domain discrepancy minimization.
Moreover, VAT also caused degenerative effect while our perturbation scheme significantly enhanced the performance, which demonstrates that conventional adversarial perturbation methods are not suitable for the SSDA problem, and our perturbation scheme can address it by modulating the perturbation direction toward the intermediate region of the target subdistributions.
The exploration scheme also worked complementary to other schemes.

\begin{figure*}[t]
    \centering
    \begin{subfigure}[b]{0.32\textwidth}
        \includegraphics[width=\textwidth]{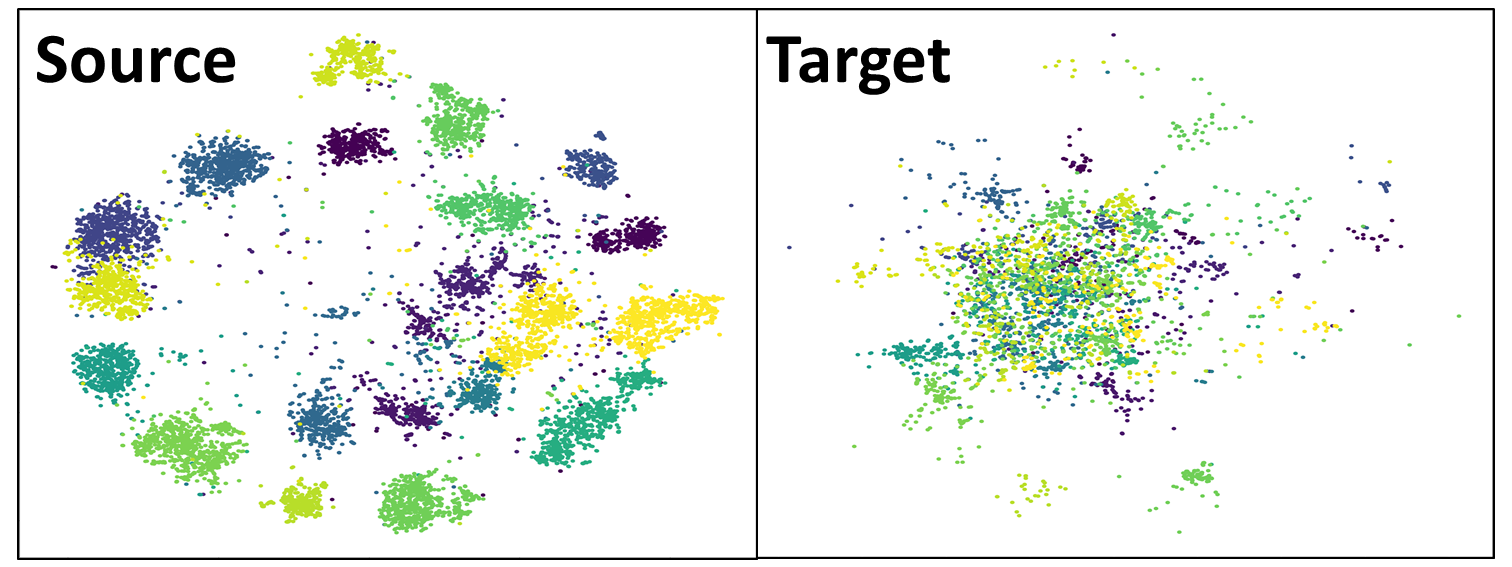}
        \caption{Iteration: 0}
    \end{subfigure}
    \begin{subfigure}[b]{0.32\textwidth}
        \includegraphics[width=\textwidth]{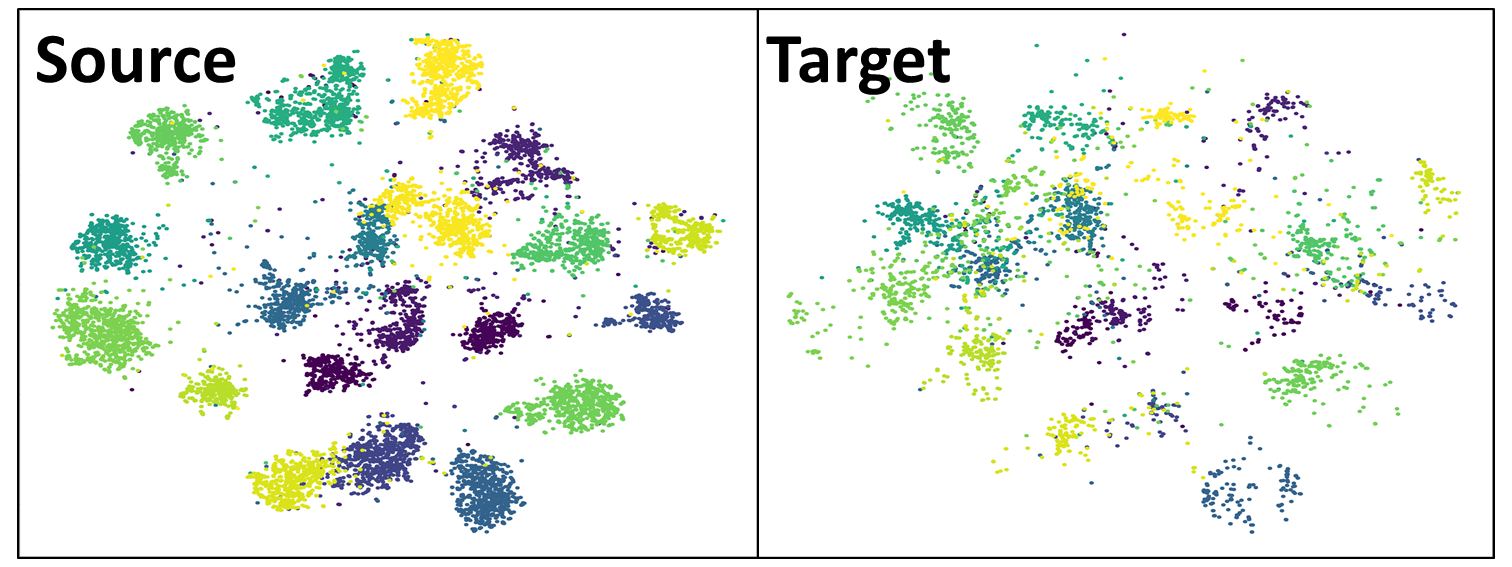}
        \caption{Iteration: 1k}
    \end{subfigure}
    \begin{subfigure}[b]{0.32\textwidth}
        \includegraphics[width=\textwidth]{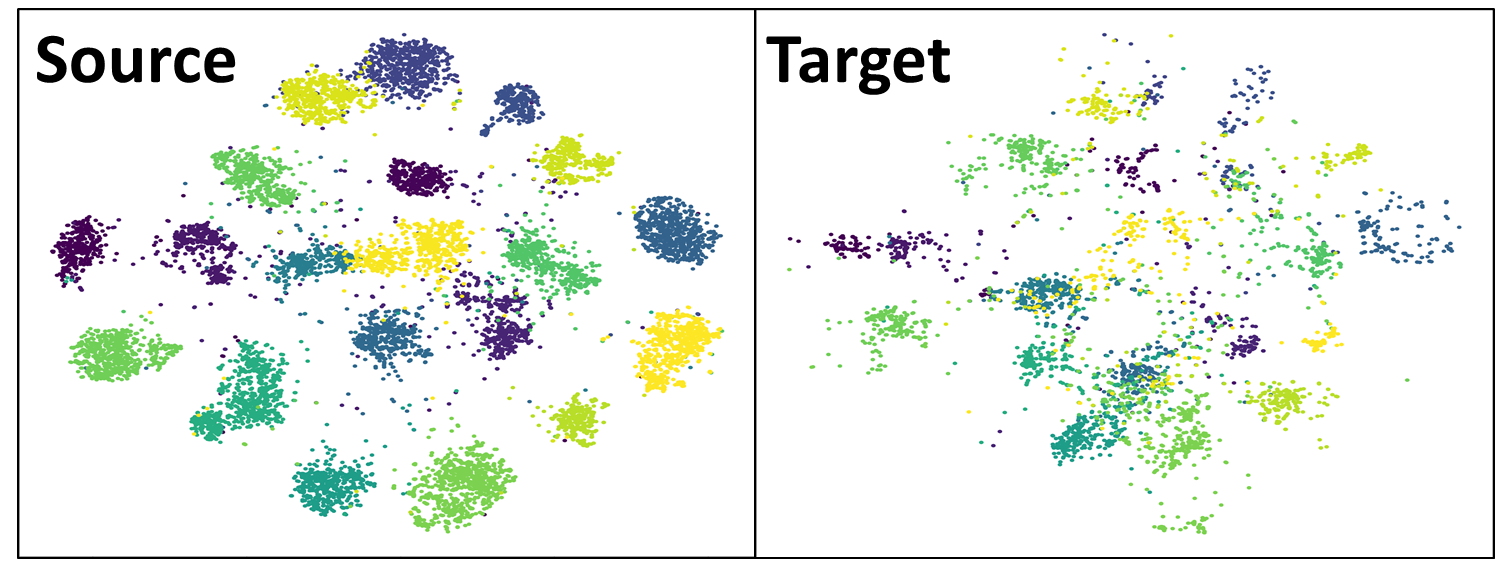}
        \caption{Iteration: 2k}
    \end{subfigure}
 
        \begin{subfigure}[b]{0.32\textwidth}
        \includegraphics[width=\textwidth]{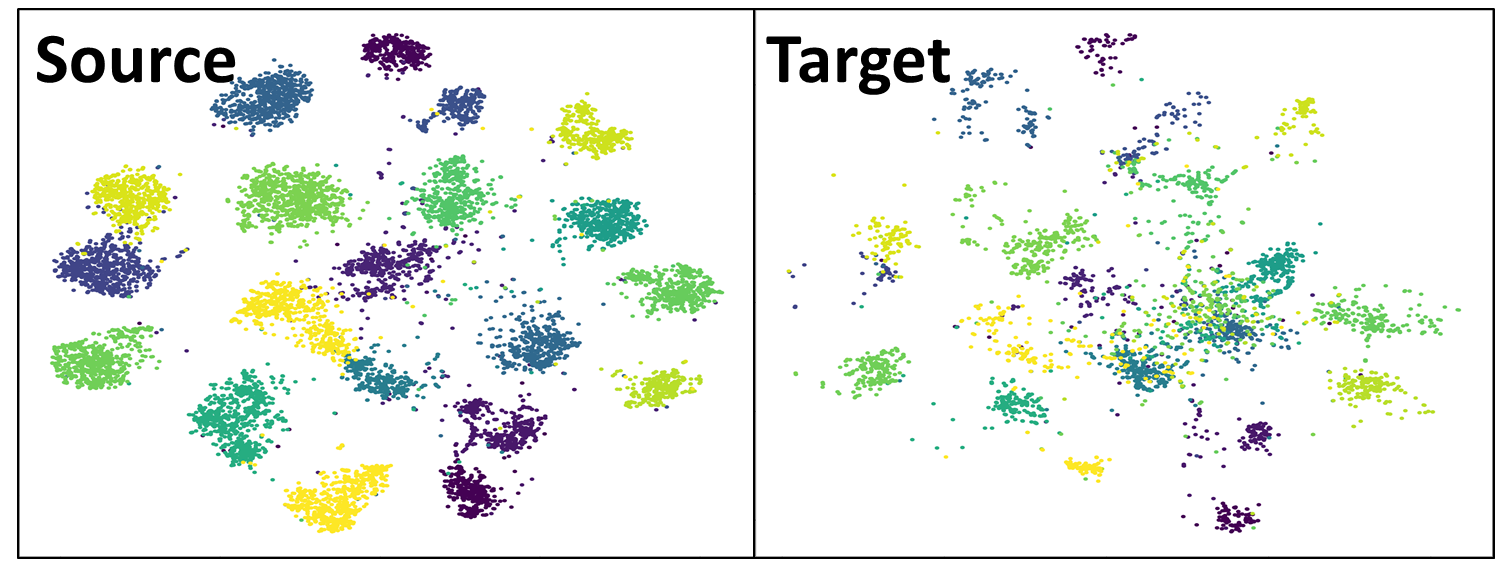}
        \caption{Iteration: 5k}
    \end{subfigure}
    \begin{subfigure}[b]{0.32\textwidth}
        \includegraphics[width=\textwidth]{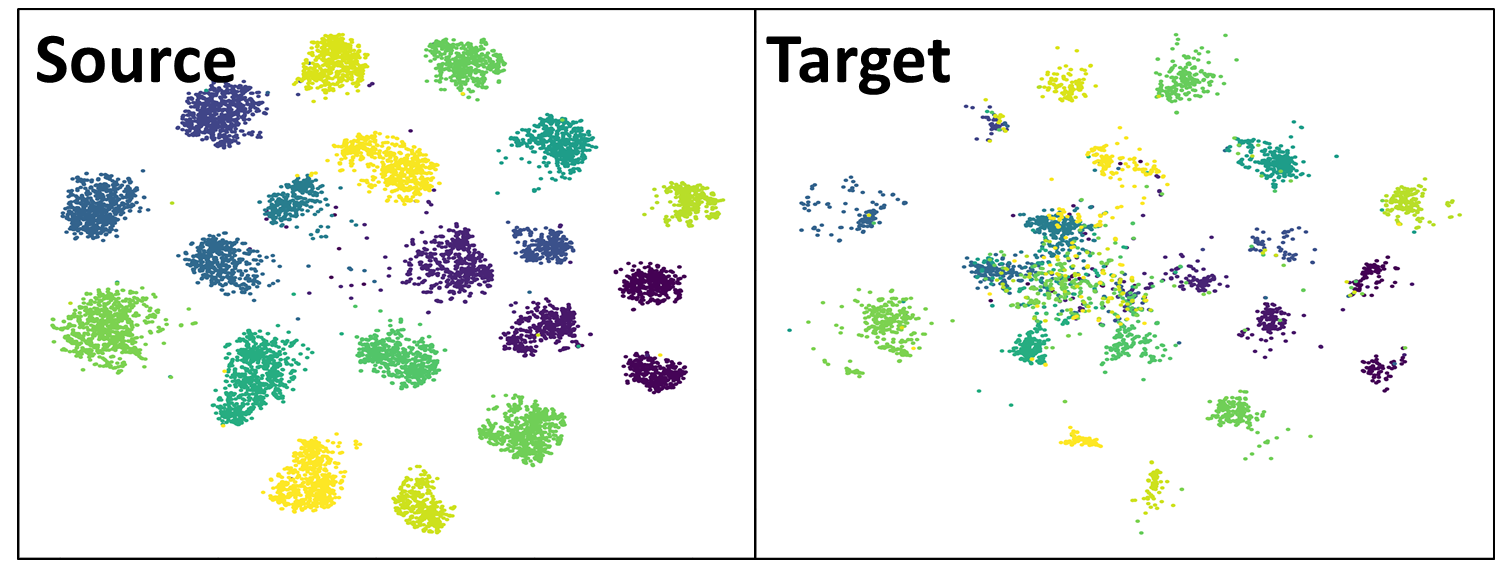}
        \caption{Iteration: 30k}
    \end{subfigure}
    \begin{subfigure}[b]{0.32\textwidth}
        \includegraphics[width=\textwidth]{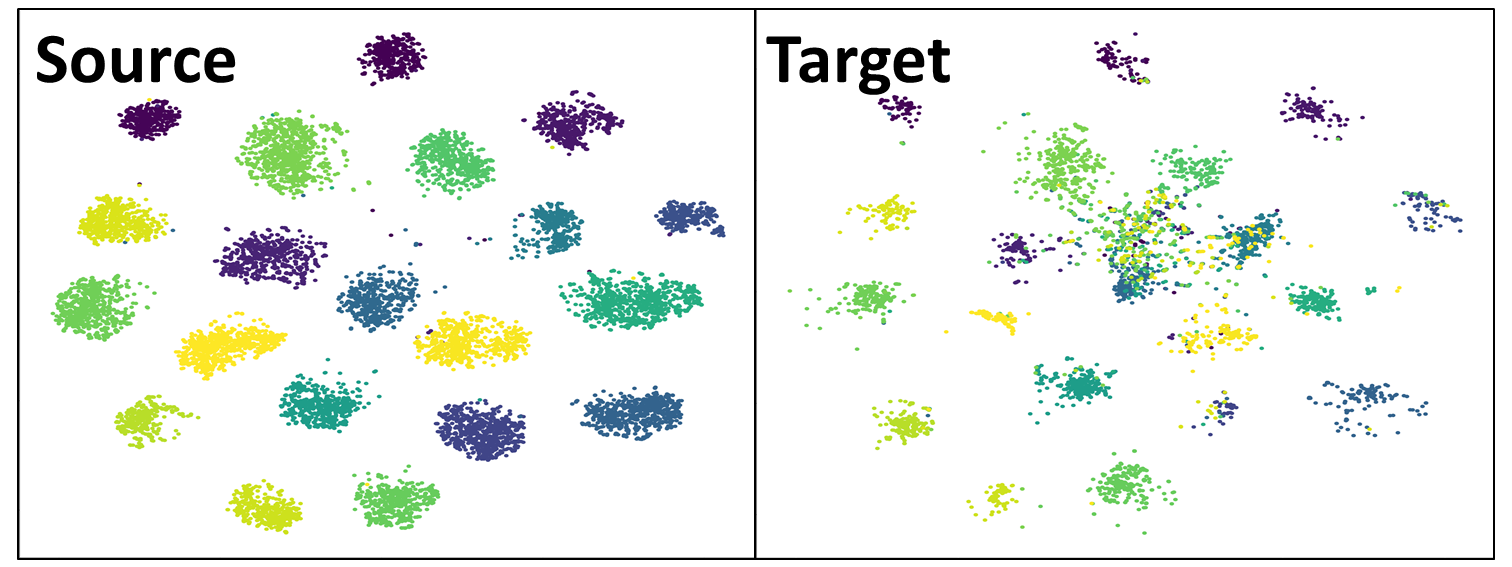}
        \caption{Iteration: 70k}
    \end{subfigure}
    
    \caption{(a)-(f) The t-SNE visualization of the feature alignment progress through our method during the training phase.
    }
    \label{fig:analysis_conv}
\end{figure*}

\noindent{\bf Convergence Analysis.} To analyze the convergence of our method, we depicted the t-SNE visualization~\cite{vanDerMaaten2008} of the cross-domain features over the training progress in Fig.~\ref{fig:analysis_conv}. 
We conducted the experiment on the Real to Sketch scenario of the DomainNet.
All 126 classes were used for the experiment, but we choose 20 classes for better visualization.
Note that we did not specifically pick classes of top-20 accuracies.
Figure~\ref{fig:analysis_conv}~(a) clearly shows the initial domain divergence between the source and the target domain.
Moreover, the feature depiction of the early stages often showed many unaligned source and target clusters.
As the training goes on, our method aligned the corresponding source and target clusters and finally obtained well-accumulated target clusters, as shown in Fig.~\ref{fig:analysis_conv}~(f).

\section{Conclusions}
In this work, we demonstrated the intra-domain discrepancy issue of the target domain in the SSDA problem.
Motivated by this, we proposed an SSDA framework that aligns the cross-domain feature distributions by addressing the intra-domain discrepancy through the attraction, exploration, and perturbation schemes.
The attraction scheme directly minimized the estimated intra-domain discrepancy within the target domain 
The perturbation scheme perturbed the well-aligned and unaligned target features into the intermediate region of the target subdistributions.
The exploration scheme locally aligned features in a selective and class-wise manner complementary to the attraction and perturbation schemes.
The experiments conducted on DomainNet, Office-Home, and Office datasets validate the effectiveness of our method, and it outperformed the conventional UDA and SSL methods on all the datasets.

\clearpage
%
%
\bibliographystyle{splncs04}
\bibliography{egbib}

\clearpage
\title{Attract, Perturb, and Explore: Learning a Feature Alignment Network for Semi-supervised Domain Adaptation (Supplementary Material)} 

\titlerunning{Learning Feature Alignment for Semi-supervised Domain Adaptation}
%
\author{Taekyung Kim\orcidID{0000-0001-7401-098X} \and
Changick Kim}
\authorrunning{Taekyung Kim and Changick Kim}
%
\institute{Korea Advanced Institute of Science and Technology, Daejeon, South Korea\\
\email{\{tkkim93, changick\}@kaist.ac.kr}}

\maketitle

\begin{table*}[!htb]
\caption{Classification accuracy ($\%$) on the DomainNet dataset with the ResNet-34 backbone network. 
The performance comparisons were done for 7 scenarios on five-shot and ten-shot settings.}
\begin{center}
\scalebox{1.0}{
\begin{tabular}{c|l|ccccccc|c}
\toprule[1.5pt] 
Net& Method       &R to C& R to P & P to C & C to S & S to P & R to S & P to R & MEAN \\\hline

\multicolumn{10}{c}{\bf Five-shot}\\\hline
\multirow{7}{*}{ResNet} & S+T          & 64.5 &63.1 & 64.2 & 59.2 & 60.4 & 56.2 & 75.7 & 63.3 \\
& DANN         & 63.7 & 62.9 & 60.5 & 55.0 & 59.5 & 55.8 & 72.6 & 61.4 \\
& CDAN         & 68.0 & 65.0 & 65.5 & 58.0 & 62.8 & 58.4 & 74.8 & 64.6\\
& ENT          & 77.1 & 71.0 & 75.7 & 61.9 & 66.2 & 64.6 & {\bf 81.1} & 71.1\\
& MME  & 75.5 & 70.4 & 74.0 & 65.0 & 68.2 & 65.5 & 79.9 & 71.2\\
& Ours & {\bf 77.7} & {\bf 73.0} & {\bf 76.9} & {\bf 67.0} & {\bf 71.4} & {\bf 68.8} & 80.5 & {\bf 73.6}\\\hline

\multicolumn{10}{c}{\bf Ten-shot}\\\hline
\multirow{7}{*}{ResNet} & S+T          & 68.5 & 66.4 & 69.2 & 64.8 & 64.2 & 60.7 & 77.3 & 67.3\\
& DANN         & 70.0 & 64.5 & 64.0 & 56.9 & 60.7 & 60.5 & 75.9 & 64.6 \\
& CDAN         & 69.3 & 65.3 & 64.6 & 57.5 & 61.6 & 60.2 & 77.0 & 65.1\\
& ENT          & 79.0 & 72.9	& 78.0	& 68.9	& 68.4	& 68.1	& 82.6 & 74.0
\\
& MME & 77.1 & 71.9	& 76.3	& 67.0	& 69.7	& 67.8 & 81.2 & 73.0\\
& Ours & {\bf 79.8} & {\bf 75.1} & {\bf 78.9}  & {\bf 70.5} & {\bf 73.6} & {\bf 70.8} & {\bf 82.9} & {\bf 76.8}\\
        \bottomrule[1.5pt]

\end{tabular}}
\end{center}
\label{tb:varying2}
\end{table*}

\section{Additional Results and Analysis}
{\bf Additional Performance Comparisons with Varying Number of Target Labels.}
We additionally conducted performance comparisons on five-shot and ten-shot settings for all the scenarios in the DomainNet dataset.
We used ResNet-34 as the backbone network for the experiments.
As shown in Table~\ref{tb:varying2}, our method outperformed all the baselines with a considerable margin on average on both the five-shot and ten-shot settings.
Moreover, our method showed superior results on all the scenarios compare to the other baselines except only one scenario, which verifies the superiority of our method on conventional domain adaptation methods, conventional semi-supervised methods, and the state-of-the-art semi-supervised domain adaptation method.
On the other hand, while the conventional domain adaptation methods showed less effective or even worse performances than S+T, ENT showed comparable results on the five-shot setting and outperformed the state-of-the-art SSDA method on the ten-shot setting on average.
As discussed in the main paper, it is because the SSDA problem gradually resembles the SSL problem on the target domain as the number of labeled target samples increases.
Note that the five-shot and ten-shot settings stand for 1.8$\%$ $\sim$ 6.7$\%$ and 3.6$\%$ $\sim$ 13.5$\%$ ratio among all target samples on the target domains of the DomainNet dataset.

\end{document}